\documentclass[a4paper,twoside]{article}
\usepackage{epsfig}
\usepackage{subcaption}
\usepackage{float}
\usepackage{calc}
\usepackage{amssymb}
\usepackage{amstext}
\usepackage{amsmath}
\usepackage{amsthm}
\usepackage{multicol}
\usepackage{pslatex}
\usepackage{apalike}
\usepackage[ruled,vlined]{algorithm2e}
\usepackage{float}  
\usepackage{booktabs}
\usepackage{dblfloatfix} 
\usepackage[utf8]{inputenc}

\usepackage[bottom]{footmisc}
\usepackage{SCITEPRESS}     

\begin{document}

\title{Robust Optical Flow Computation: A Higher-Order Differential Approach}

\author{\authorname{Chanuka Algama\sup{1}\orcidAuthor{0009-0009-3808-2869}, Kasun Amarasinghe\sup{2}\orcidAuthor{0000-0001-5143-3031} }
\affiliation{\sup{1}Tulane University, New Orleans, LA, USA}
\affiliation{\sup{2}Carnegie Mellon University, Pittsburgh, PA, USA}
\email{calgama@tulane.edu, kamarasi@andrew.cmu.edu}
}


\abstract{In the domain of computer vision, optical flow stands as a cornerstone for unraveling dynamic visual scenes. However, the challenge of accurately estimating optical flow under conditions of large nonlinear motion patterns still remains an open question. The image flow constraint is vulnerable to rapid spatial transformations, while inaccurate approximations inherent in numerical differentiation techniques can further amplify such intricacies. In response, this research proposes an innovative algorithm for optical flow computation, by incorporating second-order Taylor series approximation of the brightness constancy constraint, augmented with perturbation-theoretic correction scheme within a differential estimation framework. By embracing this mathematical underpinning, the research seeks to extract more information about the behavior of the function under complex real-world scenarios and estimate the motion of areas with a lack of texture. Although deep learning-based optical flow computation models have shown impressive results, their high resource demands, large training datasets, and black-box nature limit their interpretability and applicability in real-world scenarios, especially in cases not well-represented in the training data. An impressive showcase of the proposed algorithm's capabilities emerges through its performance on renowned optical flow benchmarks such as KITTI (2015) and Middlebury. The average endpoint error (AEE), which computes the Euclidian distance between the calculated flow field and the ground truth flow field, stands notably diminished, validating the effectiveness of the algorithm in handling complex motion patterns.}

\onecolumn \maketitle \normalsize \setcounter{footnote}{0} \vfill

\section{\uppercase{Introduction}}
\label{sec:introduction}

Accurate computation of optimal correspondence between pairs of points is one of the fundamental problems in computer vision. For decades, optical flow has been a primary tool in computing, for each pixel in one image, an optimal corresponding pixel in the resultant image of a small time step. It plays a pivotal role in understanding the motion dynamics within visual scenes. It is a 2D vector array where each vector is the motion of a pixel between two subsequent frames \cite{forsyth_learning_2008}. Applications range from autonomous navigation \cite{chao_survey_2014}, \cite{menze_object_2015-1}, breast tumor detection \cite{abdel-nasser_analyzing_2017}, bladder cancer detection \cite{weibel_graph_2012}, drone/video stabilization, traffic monitoring \cite{kastrinaki_survey_2003} and fluid mechanics \cite{inproceedings} etc. 

\begin{figure}[H]
    \centering
    \begin{subfigure}[b]{0.48\columnwidth}
        \centering
        \includegraphics[width=\linewidth]{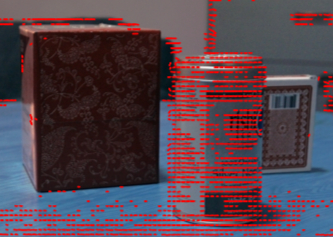}
    \end{subfigure}
    \hfill
    \begin{subfigure}[b]{0.48\columnwidth}
        \centering
        \includegraphics[width=\linewidth]{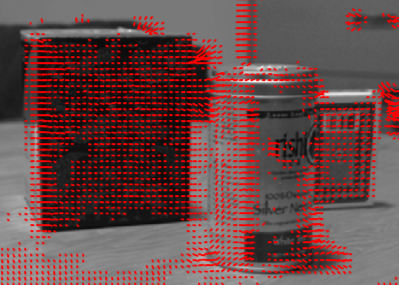}
    \end{subfigure}
    
    \begin{subfigure}[b]{0.48\columnwidth}
        \centering
        \includegraphics[width=\linewidth]{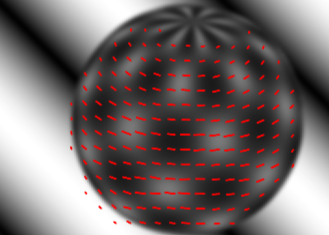}
        \caption*{(a)}
    \end{subfigure}
    \hfill
    \begin{subfigure}[b]{0.48\columnwidth}
        \centering
        \includegraphics[width=\linewidth]{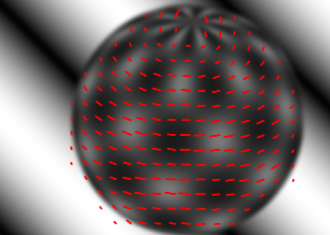}
        \caption*{(b)}
    \end{subfigure}
    
    \caption{Qualitative comparison of computed optical flow using (a) the Lucas-Kanade implementation in OpenCV using 1st order approximation and (b) the proposed algorithm. The proposed method demonstrates improved motion capturing in low-texture regions (indicated in red lines, where the direction of the lines corresponds to the directions of the motion), specifically within the box in the first example and in the upper part of the ball in the second example where the pixels lack texture.}
    \label{fig:comparison}
\end{figure}

Despite its broad application, reliable computation of flow fields under complex real-world scenarios is a major issue. Particularly when dealing with large non-linear motion patterns, rapid spatial transformations, and regions lacking texture, often resulting in inaccurate flow computations. Over the past two decades, numerous techniques and novel concepts have evolved to overcome this. In particular, the past decade has seen extraordinary improvement with the development of advanced computer vision benchmarks suites like KITTI \cite{geiger_are_2012}, Middlebury \cite{baker_database_2011}, and MPI-Sintel \cite{butler_naturalistic_2012} which provided the algorithms with significant novel hurdles in real-world settings. 

Although deep learning-based optical flow computation models such as FlowNet \cite{flownet}, PWC-Net \cite{inproceedingspwcNet} achieve impressive results, their need for high compute power and large number of training samples limits their deployment in the real world. Moreover, the black-box nature of these models makes it difficult to interpret their behavior in complex scenarios. Their performance often degrades under domain shifts---when tested in scenes that differ from their training distributions---and the scarcity of reliable ground-truth flow annotations in natural settings exacerbates this issue.

Existing computation methods are largely based on differential approach, which involves using spatial and temporal derivatives of the image intensities to compute the flow \cite{verri1990differential}, while block matching methods, energy-based methods, correlation-based methods and learning-based methods are other prominent computation approaches \cite{liu2017block} \cite{barron1994performance}.

Despite decades of research, a fundamental limitation of most classical methods is the assumption of small, linear motion between consecutive frames. Due to this, in practice the basic flow constraint tends to break down, leading to significant errors. The pyramidal (multi-scale) implementations are also subject to this faltering, since the linear approximation at each scale may not fully capture the true transformation \cite{barnum2003exploring}.
Another key source of error is the sensitivity of gradient-based methods to noise from the spatial and temporal intensity gradients. Especially in low-frame rate settings, small measurement errors in derivatives can be greatly amplified \cite{barnum2003exploring}.
Occlusions and motion discontinuities also break the flow computation. A more robust algorithm would be a one that can handle large inter-frame motion (or low frame rates) and exploit temporal coherence over multiple frames. In effect, even if an object is hidden for a few frames, the algorithm can still recover the correct correspondence in the later frame by leveraging motion priors, and temporal regularization.

In texture-poor regions (e.g. blank walls or clear sky), image gradients are weak or nearly uniform, and the aperture problem becomes severe. Most of the current computation methods neglect higher-order terms when linerizing the brightness constancy constraint. However, the impact of it under texture-poor conditions has not been systematically explored.

Therefore in this study, our key contribution is to transform optical flow computation into a formulation that:

\begin{itemize}
    \item is robust to non-linear motion patterns and rapid spatial transformations.
    \item achieves greater stability in motion estimation across texture-poor regions, where weak gradients and the aperture problem typically degrade accuracy.
    \item offers an easy plug-and-play mechanism, for seamless integration into existing optical flow pipelines and downstream vision applications.
    \item preserves linear time profile of classical differential flow, even after combining higher order polynomials into the function.
\end{itemize}

\section{\uppercase{Related Work}}

Here we discuss flow estimation methods that are closely related to our work and their limitations.

\subsection{Classical Approaches}

Early works such as Horn and Schunck (HS) global method \cite{horn1981determining} and Lucas and Kanade (LK) local differential approach laid the basis for the computation of optical flow. They rely on the assumption of brightness constancy and small inter-frame displacements, with HS enforcing a smoothness prior on the flow field, and LK solving for flow in small neighborhoods via least squares, which leads to the following derivation:

Let $I(x,y,t)$ be the function of intensity at position $(x,y)$ and time $t$. Assuming constant intensity across two consecutive frames, we have:

$$
I(x,y,t) = I(x+u,y+v,t+dt)
$$
Using Taylor series expansion around $(x,y,t)$, we approximate the right-hand side of the equation:

$$
\begin{aligned}
I(x+u,y+v,t+dt) &= I(x,y,t) + dx\frac{\partial I}{\partial x} + dy\frac{\partial I}{\partial y} + dt\frac{\partial I}{\partial t} \\
&+ \text{higher-order terms}
\end{aligned}
$$
Subtracting $I(x,y,t)$ from both sides and ignoring higher-order terms gives us:
$$
dx\frac{\partial I}{\partial x} + dy\frac{\partial I}{\partial y} + dt\frac{\partial I}{\partial t} = 0
$$
Dividing by $dt$, we obtain the fundamental optical flow constraint equation:
$$
\frac{\partial I}{\partial x}u + \frac{\partial I}{\partial y}v + \frac{\partial I}{\partial t} = 0
$$
Where $u$ and $v$ represent the horizontal and vertical components of the optical flow velocity, and $dt$ is the time difference between frames. $\frac{\partial I}{\partial x}$, $\frac{\partial I}{\partial y}$, and $\frac{\partial I}{\partial t}$ denote the spatial and temporal derivatives of the image intensity function $I(x,y,t)$.

Over time, researchers introduced improvements to relax these assumptions and handle more complex scenarios. For example, multiscale pyramid schemes were proposed to cope with larger motions by iteratively warping images and refining flow estimates at increasing resolutions \cite{article_dan}. However, even coarse-to-fine strategies can break down when motion between frames is highly non-linear or the initial guess at a coarser level is poor \cite{article_dan}. \cite{lai_no_1998} proposed two efficient algorithms for optical flow -- including a modified gradient-based regularization method -- to overcome errors caused by numerical differentiation techniques in the image flow constraint.

\begin{algorithm}[!t]
\caption{Proposed Algorithm}
\label{alg:second_order_of_alg2e}
\KwIn{Frames $I_1, I_2$; levels $L$; scale $s\in(0,1)$; iterations $K$; window $w$; derivative smoothing $\sigma$; ridge $\gamma$; correction iters $\texttt{corr\_iters}$}
\KwOut{Dense flow $(u,v)$}

\SetKwFunction{BuildPyr}{BuildPyramid}
\SetKwFunction{Warp}{Warp}
\SetKwFunction{Derivs}{Derivatives}
\SetKwFunction{WinSums}{WindowSums}
\SetKwFunction{Solve}{Solve2x2}

$\{I_1^{(\ell)}\}_{\ell=1}^L \gets \BuildPyr(I_1)$,\;

$\{I_2^{(\ell)}\}_{\ell=1}^L \gets \BuildPyr(I_2)$ 
\BlankLine
Initialize $u\gets 0$, $v\gets 0$ at level $\ell{=}1$ resolution\;

\For{$\ell \gets 1$ \KwTo $L$}{%
  \If{$\ell>1$}{Upsample $(u,v)$ to resolution of $I_1^{(\ell)}$ and scale by $1/s$\;}
  
  \For{$k \gets 1$ \KwTo $K$}{%
    $I_2^{w} \gets \Warp(I_2^{(\ell)}, u, v)$ \tcp{backward warping}
    $(I_x, I_y, I_t, I_{xx}, I_{xy}, I_{yy}, I_{xt}, I_{yt}) \gets \Derivs(I_1^{(\ell)}, I_2^{w}, \sigma)$ \tcp{$I_{tt}\approx 0$}
    $(S_{xx}, S_{xy}, S_{yy}, S_{xt}, S_{yt}) \gets \WinSums(I_x^2, I_xI_y, I_y^2, I_xI_t, I_yI_t, w)$\;
    \BlankLine
    $G \gets \begin{bmatrix} S_{xx}{+}\gamma & S_{xy} \\ S_{xy} & S_{yy}{+}\gamma \end{bmatrix}$,\quad
    
    $b \gets -\begin{bmatrix} S_{xt} \\ S_{yt} \end{bmatrix}$\;
    $(\Delta u_0,\Delta v_0) \gets \Solve(G,b)$ \tcp*{first-order LK increment}
    
    $(\tilde u,\tilde v) \gets (u{+}\Delta u_0,\, v{+}\Delta v_0)$\;

    \For{$c \gets 1$ \KwTo $\texttt{corr\_iters}$}{%
      $I_2^{w} \gets \Warp(I_2^{(\ell)}, \tilde u, \tilde v)$\;
      $(I_x, I_y, I_t, I_{xx}, I_{xy}, I_{yy}, I_{xt}, I_{yt}) \gets \Derivs(I_1^{(\ell)}, I_2^{w}, \sigma)$\;
      \(
    \begin{aligned}
    q \gets \tfrac{1}{2}\big(& I_{xx}\Delta u_0^2 + 2I_{xy}\Delta u_0\Delta v_0 + I_{yy}\Delta v_0^2 \\
                         & {}+ 2I_{xt}\Delta u_0 + 2I_{yt}\Delta v_0 \big)
    \end{aligned}
    \)\;

      $(S_{xq}, S_{yq}) \gets \WinSums(I_x q, I_y q, w)$\;
      Solve $G \begin{bmatrix}\Delta u_1\\ \Delta v_1\end{bmatrix} = - \begin{bmatrix} S_{xq}\\ S_{yq}\end{bmatrix}$ via \Solve{}\;
      $(\Delta u_0,\Delta v_0) \gets (\Delta u_0{+}\Delta u_1,\, \Delta v_0{+}\Delta v_1)$\;
      $(\tilde u,\tilde v) \gets (u{+}\Delta u_0,\, v{+}\Delta v_0)$\;
    }
    $(u,v) \gets (\tilde u,\tilde v)$\;
  }
}
\Return $(u,v)$\;
\end{algorithm}

\subsection{Higher-Order Models and Second-Order Approaches}

To address the limitations of the classical brightness constancy model, researchers explored incorporating higher-order information into the estimation. \cite{papenberg2006highly} explored higher order polynomial constancy like laplacian of the image and the determinant as a hessian, by formulating the computation of optical flow as an energy minimization problem. \cite{black1996robust} questioned that even though higher-order polynomials can model complex situations, they can still be very smooth, which can allow for more motion occlusions. For example, the 8-parameter quadratic model extends the affine motion model to represent the rigid motion of a planar surface under perspective projection \cite{fortun2015optical}. In contrast, our approach presented in this study differs conceptually: rather than explicitly modeling the motion field through higher-order polynomials, it extends the differential formulation of the brightness constancy constraint augmented with the perturbation-theoretic correction scheme without enforcing global smoothness.

Moreover, penalizing higher-order derivatives of the flow alleviates the well-known staircasing artifact where the flow in low-texture regions becomes piecewise constant. \cite{trobin2008unbiased} introduced an unbiased smoothness prior that penalizes the curvature of the flow field rather than its gradient. They discussed how higher-order regularizer greatly reduced the bias toward zero flow in weakly-textured areas \cite{fortun2015optical}. 

These approaches, whether focusing in the data constraint or smoothness term, differ from our proposed method: they do not explicitly incorporate an expansion of the brightness constancy constraint itself. Instead, they either add auxiliary invariants or improve smoothness of the solution. A closer attempt to truly utilize a differential model can be found in \cite{heitz2008dynamic} where they derived quantiles like vorticity and strain directly from velocity gradients in particle image velocimetry and fluid flow consistency, rather than general two-frame optical flow in the wild.

Additionally, performance of these studies with state-of-the-art benchmarks under real-world conditions, particularly in sequences with moderate or low frame rates that induce large inter-frame displacements, remains unexplored. 

\subsection{Learning Based Approaches}

In the past decade, CNN-based methods such as FlowNet \cite{flownet} and PWC-Net \cite{inproceedingspwcNet} achieved unprecedented precision by learning to regress flow from large labeled datasets, significantly outperforming classical variational methods on benchmarks. More recently, architectures such as RAFT (Recurrent All-Pairs Field Transforms) further improved accuracy by iteratively refining a dense flow field with learned update operators \cite{teed2020raft}. Despite their success, the requirement for larger training datasets, extensive computation which hinders the ability to be deployed in resource-constrained real-time settings, the tenancy to be specialized in training data distribution and lack of interpretability in critical safety applications motivate continued interest in classical and hybrid methods \cite{zhai2021optical}.

\section{\uppercase{Proposed Algorithm}}

\subsection{Mathematical Derivation}

Building block for our derivation is the classical brightness constancy assumption, which states that the intensity of a pixel remains constant between two consecutive frames.

\begin{equation}
I(x+u,\, y+v,\, t+\Delta t) = I(x,\, y,\, t),
\label{eq:brightness_constancy}
\end{equation}

where $I(x,y,t)$ denotes the intensity of the image in the pixel $(x,y)$ and the time $t$, and $(u,v)$ represents the displacement vector of that pixel over a short time interval $\Delta t$. 

Defining the residual brightness difference as
\begin{equation}
r(u,v) = I(x+u, y+v, t+\Delta t) - I(x, y, t) = 0,
\end{equation}

In general, the Taylor series for a function $f(x)$ is:

\begin{align}
\sum_{n=0}^{\infty} \frac{f^{(n)}(a)}{n!}(x-a)^n &= f(a) + f'(a)(x-a) + \nonumber \\
\frac{f''(a)}{2!}(x-a)^2
&\quad + \dots + \frac{f^{(k)}(a)}{k!}(x-a)^k + \dots
\nonumber
\end {align}

For the multi variable intensity function, the linear approximation is given by the first-order Taylor expansion:

\begin{align}
F(\overline{x}) \equiv f(\overline{a}) + D(\overline{a}) (\overline{x} - \overline{a})
\nonumber
\end{align}

where $Df(\overline{a})$ is the Jacobian matrix, representing the first-order expansion evaluated at point $\overline{a}$. Expanding $I(x+u, y+v, t+\Delta t)$ around $(x, y, t)$ using a second-order Taylor expansion using hessian matrix yields

\begin{align}
I(x+u, y+v, t+\Delta t) \approx\; & I(x,y,t)
+ I_x u + I_y v + I_t \Delta t \nonumber \\ 
& + \frac{1}{2} \big( I_{xx} u^2 + 2 I_{xy} uv + I_{yy} v^2 \nonumber \\
& + 2 I_{xt} u \Delta t + 2 I_{yt} v \Delta t + I_{tt} \Delta t^2 \big),
\end{align}

where subscripts denote partial derivatives of $I$ with respect to the corresponding variables. Substituting into (\ref{eq:brightness_constancy}) and cancelling $I(x,y,t)$ gives the \emph{second-order brightness constancy constraint}:

\begin{align}
I_x u + I_y v + I_t \Delta t
+ \frac{1}{2}\big( I_{xx} u^2 + 2 I_{xy} uv + I_{yy} v^2 \nonumber \\
+ 2 I_{xt} u \Delta t + 2 I_{yt} v \Delta t + I_{tt} \Delta t^2 \big) = 0.
\label{eq:second_order_constraint}
\end{align}

For consecutive image frames, $\Delta t=1$ can be assumed without loss of generality, leading to

\begin{align}
I_x u + I_y v + I_t
+ \frac{1}{2} \big( I_{xx} u^2 + 2 I_{xy} uv + I_{yy} v^2 \nonumber \\
+ 2 I_{xt} u + 2 I_{yt} v + I_{tt} \big) = 0.
\label{eq:bc2}
\end{align}

The inclusion of second-order spatial and temporal derivatives allows the model to capture curvature and non-linear motion patterns that are not represented in the classical linearized formulation. When $I_x, I_y$ are weak (texture-poor), the pure $I_x u + I_y$ terms are under-constrains flow. Second-order curvature terms $I_{xx}, I_{xy}, I_{yy}$ and cross space--time terms $I_{xt}, I_{yt}$ carry extra information.

\subsection{Perturbative Correction Formulation}

Since Equation~(\ref{eq:bc2}) is non-linear in $(u,v)$, a robust two stage perturbation linearization is applied. 

\subsubsection{Stage A: First Order Estimate}

We can compute a standard first flow $(u_0, v_0)$ using LK (local least squares over a window $w$) \cite{bruhn2005lucas} or HS (variational) \cite{horn1981determining}. For LK the normal equation over window $w$ are:

\begin{equation}
\underbrace{
\sum_{\mathcal{W}}
\begin{bmatrix}
I_x^2 & I_x I_y\\
I_x I_y & I_y^2
\end{bmatrix}}_{G}
\begin{bmatrix} u_0 \\ v_0 \end{bmatrix}
=
-\underbrace{
\sum_{\mathcal{W}}
\begin{bmatrix}
I_x I_t \\ I_y I_t
\end{bmatrix}}_{\mathbf{b}}.
\label{eq:lk-first-order-compact}
\end{equation}

\subsubsection{Stage B: Second-order Correction}

Let the true flow be represented as

\begin{equation}
(u,v) = (u_0,v_0) + \varepsilon (u_1,v_1),
\end{equation}

where $(u_0,v_0)$ is the first-order estimate obtained from a standard LK or HS formulation, while $(u_1,v_1)$ being a small correction. 

Substituting this expression into (\ref{eq:bc2}) and expanding to first order in $\varepsilon$ yields
\begin{equation}
I_x u_1 + I_y v_1 = - q(x,y),
\label{eq:perturb_correction}
\end{equation}
where the second-order remainder $q(x,y)$ at $(u_0,v_0)$ is defined as
\begin{align}
q(x,y) = \frac{1}{2}\big( I_{xx} u_0^2 + 2 I_{xy} u_0 v_0 + I_{yy} v_0^2 \nonumber \\
+ 2 I_{xt} u_0 + 2 I_{yt} v_0 + I_{tt} \big).
\label{eq:remainder_term}
\end{align}

Now expand $r(u_0 + \varepsilon u_1, v_0 + \varepsilon v_1)$ to first order in $\varepsilon$, keeping only the Jacobian of the first-order part (this is the perturbation simplification that keep the correct linear and well-posed):

\begin{equation}
I_x (\varepsilon u_1) + I_y (\varepsilon v_1)
+ \underbrace{[\, I_x u_0 + I_y v_0 + I_t \,]}_{\text{first-order residual}}
+ q \;\approx\; 0,
\label{eq:perturbation_expansion}
\end{equation}

Since $(u_0, v_0)$ was obtained by minimizing the first order residual, we treat $[\, I_x u_0 + I_y v_0 + I_t \,] \approx 0$. Discarding higher-order $\varepsilon$ -terms from $q$, we get the linear correction constraint:

\begin{align}
    I_x u_1 + I_y v_1 \approx - q
\end{align}

The correction $(u_1,v_1)$ is obtained over a local window $\mathcal{W}$, leading to Lucas--Kanade--implementation of normal equations:
\begin{equation}
\sum_{\mathcal{W}}
\begin{bmatrix}
I_x^2 & I_x I_y \\
I_x I_y & I_y^2
\end{bmatrix}
\begin{bmatrix} u_1 \\ v_1 \end{bmatrix}
=
-\sum_{\mathcal{W}}
\begin{bmatrix}
I_x q \\ I_y q
\end{bmatrix}.
\label{eq:lk_second_order}
\end{equation}

Finally, the corrected flow is computed as
\begin{equation}
(u,v) = (u_0,v_0) + (u_1,v_1).
\end{equation}

\subsection{Comparative Analysis of the Existing and Proposed Algorithm}

Equation~(\ref{eq:lk_second_order}) preserves the same system matrix as the classical LK method, making this correction easily integrable into existing pipelines. The second-order term $q(x,y)$ introduces curvature and cross-derivative information, improving robustness in low-texture and non-linear motion regions. From a perturbation-theoretic standpoint, this formulation represents the first-order response of the flow field to the inclusion of second-order terms in the brightness constancy constraint, yielding a mathematically interpretable and computationally efficient extension to classical optical flow. 

Compared to parametric quadratic (8-parameter) models, and other higher order approaches we discussed, our algorithm captures higher-order effects without imposing a globally smooth polynomial field, preserving motion boundaries and remaining plug-compatible with LK/HS solvers. Versus deep models (FlowNet/PWC/RAFT), it trades raw peak accuracy for interpretability, zero training data, and stronger resilience under domain shift---while incurring only a modest compute bump over classical multi-scale LK due to one or two linear correction solves per iteration.

\begin{figure*}[ht]
    \centering
    \begin{subfigure}[b]{0.3\linewidth}
        \centering
        \includegraphics[width=\linewidth]{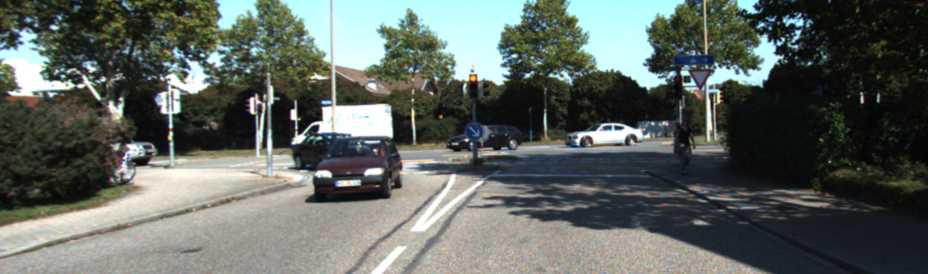}
    \end{subfigure}%
    \hspace{0.5mm}
    \begin{subfigure}[b]{0.3\linewidth}
        \centering
        \includegraphics[width=\linewidth]{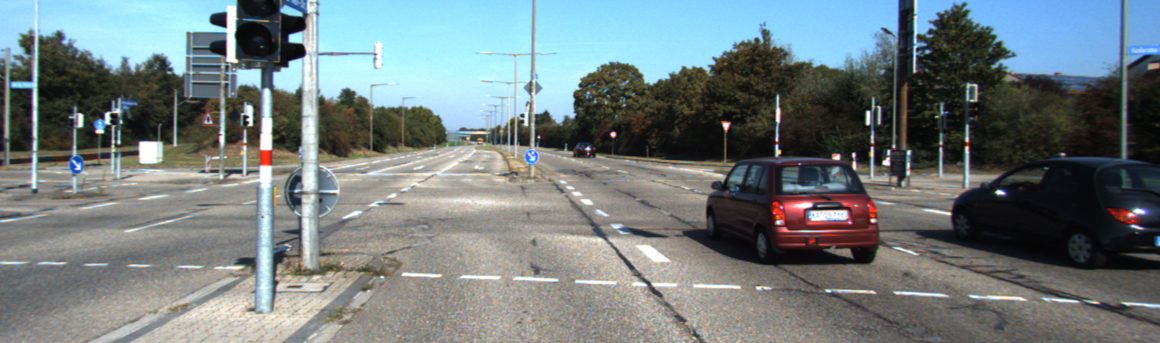}
    \end{subfigure}%
    \hspace{0.5mm}
    \begin{subfigure}[b]{0.3\linewidth}
        \centering
        \includegraphics[width=\linewidth]{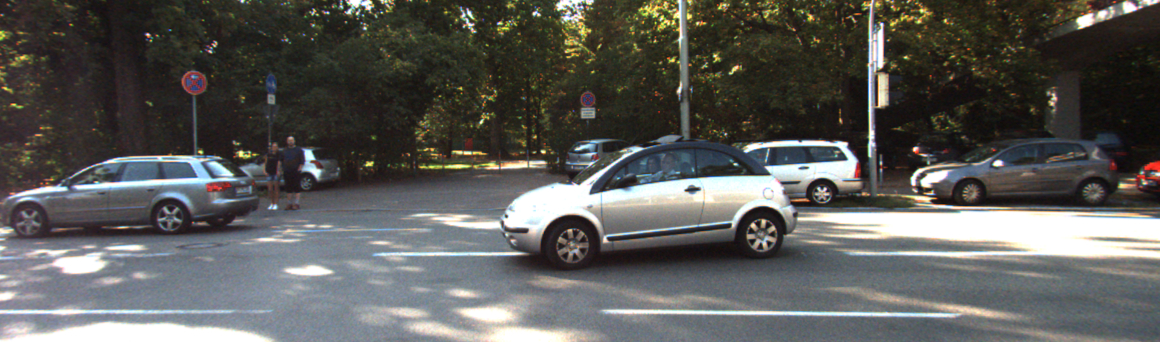}
    \end{subfigure}

    \vspace{5mm} 
    \begin{subfigure}[b]{0.3\linewidth}
        \centering
        \includegraphics[width=\linewidth]{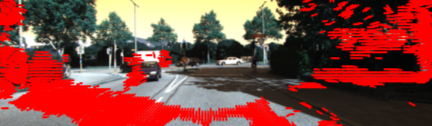}
    \end{subfigure}%
    \hspace{0.5mm}
    \begin{subfigure}[b]{0.3\linewidth}
        \centering
        \includegraphics[width=\linewidth]{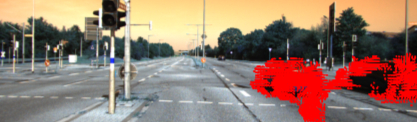}
    \end{subfigure}%
    \hspace{0.5mm}
    \begin{subfigure}[b]{0.3\linewidth}
        \centering
        \includegraphics[width=\linewidth]{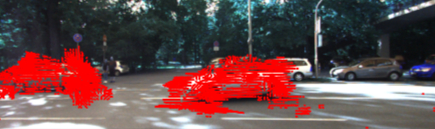}
    \end{subfigure}

    \vspace{5mm} 
    \begin{subfigure}[b]{0.3\linewidth}
        \centering
        \includegraphics[width=\linewidth]{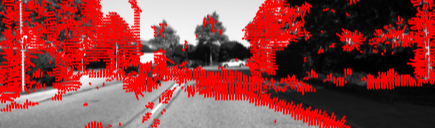}
    \end{subfigure}%
    \hspace{0.5mm}
    \begin{subfigure}[b]{0.3\linewidth}
        \centering
        \includegraphics[width=\linewidth]{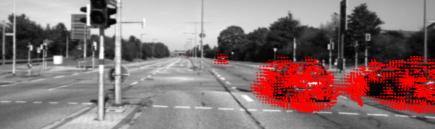}
    \end{subfigure}%
    \hspace{0.5mm}
    \begin{subfigure}[b]{0.3\linewidth}
        \centering
        \includegraphics[width=\linewidth]{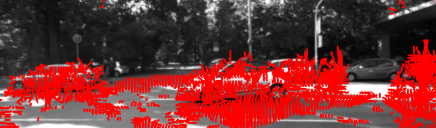}
    \end{subfigure}

    \vspace{5mm} 
    \begin{subfigure}[b]{0.4\linewidth}
        \centering
        \includegraphics[width=\linewidth]{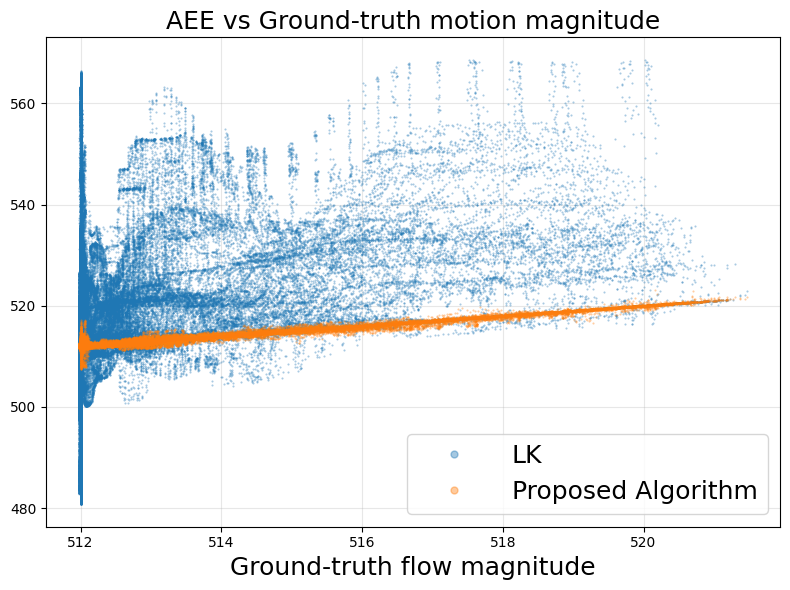}
    \end{subfigure}%
    \hspace{0.5mm}
    \begin{subfigure}[b]{0.4\linewidth}
        \centering
        \includegraphics[width=\linewidth]{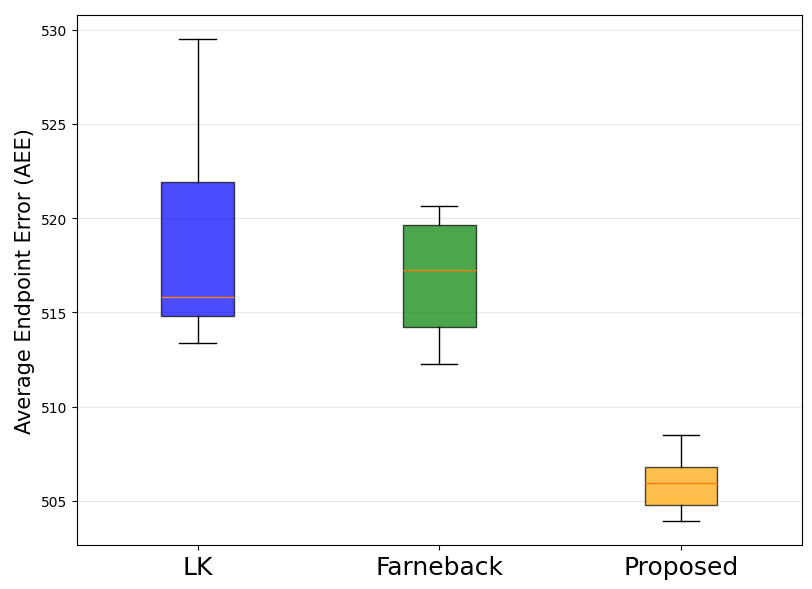}
    \end{subfigure}%

    \caption{Performance evaluation on the KITTI benchmark suite: three representative test cases showcase, the original input (first row), flow fields computed using the standard LK implementation (second row) and the proposed algorithm (3rd row). The plot (row: 4, column: 1) shows the Average Endpoint Error (AEE) as a function of the ground-truth motion magnitude. The proposed method maintains lower errors across displacement ranges and outperforms first-order Lucas--Kanade in the high-motion magnitudes, where linear approximations of brightness constancy typically break down. The plot (row: 4, column: 2) shows distribution of mean, meadian, and IQR among LK, Farnback and Proposed algorithm across 200 driving sequences in KITTI benchmark.}
    \label{kitti}
\end{figure*}

\section{\uppercase{Experimental Validations}}
\label{sec:experimental results}

Our main goal was to test the performance and robustness of the proposed algorithm under (1) large motion in between frames, (2) low-texture regions, and (3) complex non-linear real-world motion patterns. We employed the KITTI Vision Benchmark Suite \cite{geiger_are_2012}, one of the most challenging and widely adopted datasets for vision algorithm evaluation. The KITTI consists of over 200 driving sequences captured from a moving car, including range of lighting conditions, shadows, reflections and dynamic objects such as vehicles, pedestrians, and cyclists. Each scene provided along with precise, manually annotated ground-truth optical flow obtained via 3D laser scanners and stereo reconstruction. A distinctive feature of the KITTI is the substantially larger inter-frame motion (10 frames per second) compared to other benchmarks used in the studies we discussed, such as Middlebury~\cite{baker_database_2011} and MPI-Sintel~\cite{butler_naturalistic_2012}. This characteristic makes KITTI an ideal testbed for evaluating algorithms designed to handle non-linear and large-displacement motion---conditions under which traditional differential approaches often break down due to their linearization assumptions.

\begin{table*}[!t]
  \centering
  \caption{Asymptotic time complexity per image pair (dominant terms).}
  \label{tab:time_complexity}
  \begin{tabular*}{\textwidth}{@{\extracolsep{\fill}} l l l}
    \toprule
    \textbf{Method} & \textbf{Per-pair complexity} & \textbf{Notes} \\
    \midrule
    Lucas--Kanade (pyramidal) & $O\!\big(K\,\tfrac{N}{1-s^2}\big)$ & Box-filtered windows \\
    Horn--Schunck (variational) & $O\!\big(K\,\tfrac{N}{1-s^2}\big)$ & PDE iterations \\
    Farneback (quadratic local) & $O\!\big(K\,\tfrac{N}{1-s^2}\big)$ & Larger constant \\
    \textbf{Proposed (2nd-order + pert.)} & $\mathbf{O\!\big(K(1{+}C)\,\tfrac{N}{1-s^2}\big)}$ & LK matrix; small overhead \\
    PWC-Net / RAFT & $O(N)$ & High constants, GPU-friendly \\
    \bottomrule
  \end{tabular*}
\end{table*}

Figure~\ref{kitti} illustrates three representative test cases from the KITTI benchmark suite captured under diverse driving conditions. The first row presents the original input images, while the second and third rows show the flow fields computed using the standard LK implementation (first-order approximation) and the proposed algorithm, respectively. The red arrows denote the direction of the estimated motion, and their length represents the magnitude of the flow vectors. Across all scenarios of the benchmark, the proposed algorithm exhibits consistently superior performance compared to the LK and Farnback methods, achieving a substantially lower Average Endpoint Error (AEE).

To assess the algorithm's robustness in texture-deficient regions, additional experiments were conducted on image sequences containing areas with minimal spatial gradients. Figure~\ref{fig:comparison} presents two illustrative cases: (1) a tabletop scene undergoing slight rotational motion with objects placed on its surface, and (2) the motion of a rolling ball. The proposed method demonstrates significantly improved global motion capture in these challenging low-texture regions---especially within the marked box area in the first example and the upper portion of the ball in the second---where the classical LK approach fails to recover consistent motion due to insufficient gradient information.

Our method preserves the linear-time profile of classical differential flow. With efficient box filtering and warping, each outer iteration is $O(N)$ over $N{=}HW$ pixels; the second-order perturbative step adds only a constant-factor pass, yielding $T(N)=O\!\big(K(1{+}C)\,\tfrac{N}{1-s^2}\big)$ across a $L$-level pyramid with scale $s$ (typically $s{=}0.5$). In practice, $C{=}1$ suffices, so runtime is within a small multiple of multi-scale LK/HS. By contrast, global variational solvers accrue additional passes for regularization, and learning-based models run in $O(N)$ per inference but with substantial constants (feature pyramids, cost volumes).

\begin{figure}[t]
    \centering
    \includegraphics[width=\columnwidth]{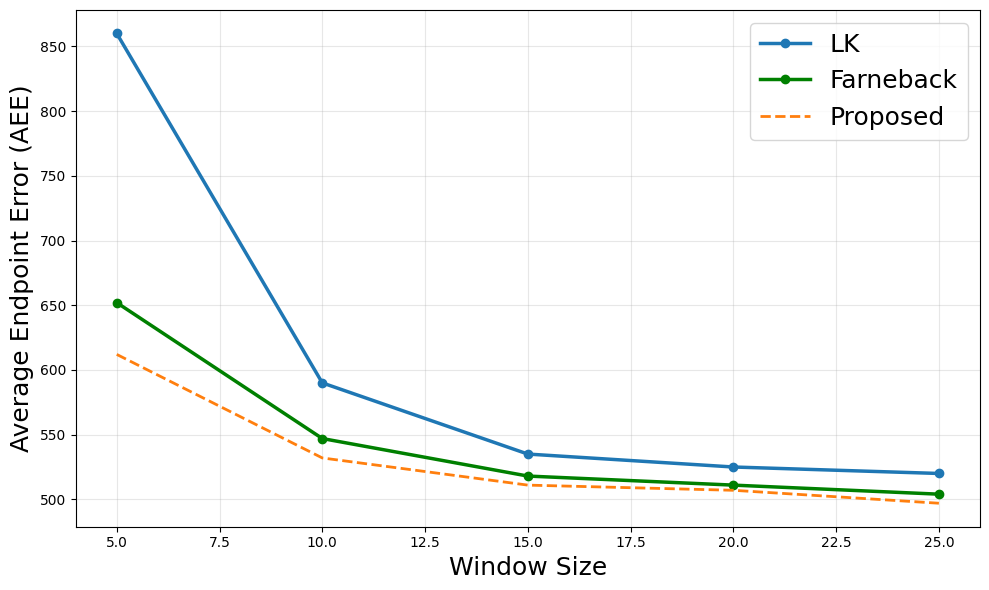}
    \caption{Average Endpoint Error (AEE) as a function of window size for Lucas--Kanade (LK), Farneback, and the proposed method. All methods exhibit decreasing error with increasing window size due to improved local motion aggregation. The proposed method consistently achieves lower AEE across window sizes, indicating better robustness to the choice of spatial support.}
    \label{fig:aee_vs_window}
\end{figure}

\section{\uppercase{Conclusions}}
\label{sec:conclusion}

In this paper, we proposed a novel algorithm for effective flow field estimation under challenging conditions, including large nonlinear motion patterns, rapid spatial transformations and estimating in areas with lack of texture. By incorporating second-order Taylor series approximation of the brightness constancy constraint, augmented with perturbation-theoretic correction scheme within a differential estimation framework. Unlike conventional first-order differential methods that assume locally linear motion, the proposed formulation explicitly models higher-order spatial and temporal variations, enabling more accurate flow estimation under non-linear and large-displacement motion.

We achieved improvements in accuracy. Rigorous testing under real-world driving scenarios, across diverse lighting conditions on KITTI 2015 benchmark characterized by low frame rate, demonstrated the robustness and reliability of the algorithm. The algorithm achieved a lower Average Endpoint Error than classical and polynomial-based baselines while retaining a compatible computational complexity. Qualitative analyses further revealed its capacity to maintain coherent flow in texture-poor regions and around motion discontinuities, where existing methods typically degrade. In summary, our work bridges a critical gap between classical analytical optical flow methods and modern real-world motion conditions. The proposed algorithm offers computationally efficient, and easily integrable solution for dense motion estimation. Future research will explore extending this framework to multi-frame temporal models and integrating it with hybrid learning-based architectures to further enhance its adaptability and real-time performance.


{\small
\begin{thebibliography}{}

\bibitem[Abdel-Nasser et~al., 2017]{abdel-nasser_analyzing_2017}
Abdel-Nasser, M., Moreno, A., A.~Rashwan, H., and Puig, D. (2017).
\newblock Analyzing the evolution of breast tumors through flow fields and
  strain tensors.
\newblock {\em Pattern Recognition Letters}, 93:162--171.

\bibitem[Baker et~al., 2011]{baker_database_2011}
Baker, S., Scharstein, D., Lewis, J.~P., Roth, S., Black, M.~J., and Szeliski,
  R. (2011).
\newblock A {Database} and {Evaluation} {Methodology} for {Optical} {Flow}.
\newblock {\em International Journal of Computer Vision}, 92(1):1--31.

\bibitem[Barnum et~al., 2003]{barnum2003exploring}
Barnum, P., Hu, B., and Brown, C. (2003).
\newblock Exploring the practical limits of optical flow.

\bibitem[Barron et~al., 1994]{barron1994performance}
Barron, J.~L., Fleet, D.~J., and Beauchemin, S.~S. (1994).
\newblock Performance of optical flow techniques.
\newblock {\em International journal of computer vision}, 12(1):43--77.

\bibitem[Beauchemin and Barron, 1995]{forsyth_learning_2008}
Beauchemin, S.~S. and Barron, J.~L. (1995).
\newblock The computation of optical flow.
\newblock {\em ACM computing surveys (CSUR)}, 27(3):433--466.

\bibitem[Black and Anandan, 1996]{black1996robust}
Black, M.~J. and Anandan, P. (1996).
\newblock The robust estimation of multiple motions: Parametric and
  piecewise-smooth flow fields.
\newblock {\em Computer vision and image understanding}, 63(1):75--104.

\bibitem[Bruhn et~al., 2005]{bruhn2005lucas}
Bruhn, A., Weickert, J., and Schn{\"o}rr, C. (2005).
\newblock Lucas/kanade meets horn/schunck: Combining local and global optic
  flow methods.
\newblock {\em International journal of computer vision}, 61(3):211--231.

\bibitem[Butler et~al., 2012]{butler_naturalistic_2012}
Butler, D.~J., Wulff, J., Stanley, G.~B., and Black, M.~J. (2012).
\newblock A {Naturalistic} {Open} {Source} {Movie} for {Optical} {Flow}
  {Evaluation}.
\newblock In Fitzgibbon, A., Lazebnik, S., Perona, P., Sato, Y., and Schmid,
  C., editors, {\em Computer {Vision} -- {ECCV} 2012}, Lecture {Notes} in
  {Computer} {Science}, pages 611--625, Berlin, Heidelberg. Springer.

\bibitem[Chao et~al., 2014]{chao_survey_2014}
Chao, H., Gu, Y., and Napolitano, M. (2014).
\newblock A {Survey} of {Optical} {Flow} {Techniques} for {Robotics}
  {Navigation} {Applications}.
\newblock {\em Journal of Intelligent \& Robotic Systems}, 73(1):361--372.

\bibitem[Fischer et~al., 2015]{flownet}
Fischer, P., Dosovitskiy, A., Ilg, E., Hausser, P., Hazirbas, C., Golkov, V.,
  van~der Smagt, P., Cremers, D., and Brox, T. (2015).
\newblock Flownet: Learning optical flow with convolutional networks.

\bibitem[Fortun et~al., 2015]{fortun2015optical}
Fortun, D., Bouthemy, P., and Kervrann, C. (2015).
\newblock Optical flow modeling and computation: A survey.
\newblock {\em Computer Vision and Image Understanding}, 134:1--21.

\bibitem[Geiger et~al., 2012]{geiger_are_2012}
Geiger, A., Lenz, P., and Urtasun, R. (2012).
\newblock Are we ready for autonomous driving? {The} {KITTI} vision benchmark
  suite.
\newblock In {\em 2012 {IEEE} {Conference} on {Computer} {Vision} and {Pattern}
  {Recognition}}, pages 3354--3361.
\newblock ISSN: 1063-6919.

\bibitem[Heitz et~al., 2008a]{article_dan}
Heitz, D., Heas, P., Memin, E., and Carlier, J. (2008a).
\newblock Dynamic consistent correlation-variational approach for robust
  optical flow estimation.
\newblock {\em Experiments in Fluids}, 45:595--608.

\bibitem[Heitz et~al., 2008b]{heitz2008dynamic}
Heitz, D., H{\'e}as, P., M{\'e}min, E., and Carlier, J. (2008b).
\newblock Dynamic consistent correlation-variational approach for robust
  optical flow estimation.
\newblock {\em Experiments in fluids}, 45(4):595--608.

\bibitem[Horn and Schunck, 1981]{horn1981determining}
Horn, B.~K. and Schunck, B.~G. (1981).
\newblock Determining optical flow.
\newblock {\em Artificial intelligence}, 17(1-3):185--203.

\bibitem[Kastrinaki et~al., 2003]{kastrinaki_survey_2003}
Kastrinaki, V., Zervakis, M., and Kalaitzakis, K. (2003).
\newblock A survey of video processing techniques for traffic applications.
\newblock {\em Image and Vision Computing}, 21(4):359--381.

\bibitem[Khalid et~al., 2017]{inproceedings}
Khalid, M., Penard, L., and Memin, E. (2017).
\newblock Application of optical flow for river velocimetry.
\newblock In {\em applications of of}, pages 6243--6246.

\bibitem[Lai and Vemuri, 1998]{lai_no_1998}
Lai, S.-H. and Vemuri, B.~C. (1998).
\newblock [{No} title found].
\newblock {\em International Journal of Computer Vision}, 29(2):87--105.

\bibitem[Liu and Delbruck, 2017]{liu2017block}
Liu, M. and Delbruck, T. (2017).
\newblock Block-matching optical flow for dynamic vision sensors: Algorithm and
  fpga implementation.
\newblock In {\em 2017 IEEE International Symposium on Circuits and Systems
  (ISCAS)}, pages 1--4. IEEE.

\bibitem[Menze and Geiger, 2015]{menze_object_2015-1}
Menze, M. and Geiger, A. (2015).
\newblock Object scene flow for autonomous vehicles.
\newblock In {\em 2015 {IEEE} {Conference} on {Computer} {Vision} and {Pattern}
  {Recognition} ({CVPR})}, pages 3061--3070, Boston, MA, USA. IEEE.

\bibitem[Papenberg et~al., 2006]{papenberg2006highly}
Papenberg, N., Bruhn, A., Brox, T., Didas, S., and Weickert, J. (2006).
\newblock Highly accurate optic flow computation with theoretically justified
  warping.
\newblock {\em International Journal of Computer Vision}, 67(2):141--158.

\bibitem[Sun et~al., 2018]{inproceedingspwcNet}
Sun, D., Yang, X., Liu, M.-Y., and Kautz, J. (2018).
\newblock Pwc-net: Cnns for optical flow using pyramid, warping, and cost
  volume.
\newblock In {\em pwc-net}, pages 8934--8943.

\bibitem[Teed and Deng, 2020]{teed2020raft}
Teed, Z. and Deng, J. (2020).
\newblock Raft: Recurrent all-pairs field transforms for optical flow.
\newblock In {\em European conference on computer vision}, pages 402--419.
  Springer.

\bibitem[Trobin et~al., 2008]{trobin2008unbiased}
Trobin, W., Pock, T., Cremers, D., and Bischof, H. (2008).
\newblock An unbiased second-order prior for high-accuracy motion estimation.
\newblock In {\em Joint Pattern Recognition Symposium}, pages 396--405.
  Springer.

\bibitem[Verri et~al., 1990]{verri1990differential}
Verri, A., Girosi, F., and Torre, V. (1990).
\newblock Differential techniques for optical flow.
\newblock {\em Journal of the Optical Society of America A}, 7(5):912--922.

\bibitem[Weibel et~al., 2012]{weibel_graph_2012}
Weibel, T., Daul, C., Wolf, D., Rosch, R., and Guillemin, F. (2012).
\newblock Graph based construction of textured large field of view mosaics for
  bladder cancer diagnosis.
\newblock {\em Pattern Recognition}, 45(12):4138--4150.

\bibitem[Zhai et~al., 2021]{zhai2021optical}
Zhai, M., Xiang, X., Lv, N., and Kong, X. (2021).
\newblock Optical flow and scene flow estimation: A survey.
\newblock {\em Pattern Recognition}, 114:107861.

\end{thebibliography}
}
\end{document}